\documentclass{article}
\usepackage{spconf,amsmath,graphicx,amsfonts}
\usepackage{booktabs}

\title{Distribution-based Emotion Recognition in Conversation}
\name{Wen Wu\thanks{Wen Wu is supported by a Cambridge International Scholarship from the Cambridge Trust. This work has been performed using resources provided by the Cambridge Tier-2 system operated by the University of Cambridge Research Computing Service (www.hpc.cam.ac.uk) funded by EPSRC Tier-2 capital grant EP/T022159/1.}, Chao Zhang, Philip C. Woodland}
\address{Department of Engineering, University of Cambridge, Trumpington St., Cambridge, UK.\\
\small{\texttt{\{ww368,cz277,pcw\}@eng.cam.ac.uk}}}
\copyrightnotice{978-1-6654-7189-3/22/\$31.00~\copyright2023 IEEE}
\begin{document}
\ninept
\maketitle
\begin{abstract}
Automatic emotion recognition in conversation (ERC) is crucial for emotion-aware conversational artificial intelligence. This paper proposes a distribution-based framework that formulates ERC as a sequence-to-sequence problem for emotion distribution estimation. The inherent ambiguity of emotions and the subjectivity of human perception lead to disagreements in emotion labels, which is handled naturally in our framework from the perspective of uncertainty estimation in emotion distributions. A Bayesian training loss is introduced to improve the uncertainty estimation by conditioning each emotional state on an utterance-specific Dirichlet prior distribution. Experimental results on the IEMOCAP dataset show that ERC outperformed the single-utterance-based system, and the proposed distribution-based ERC methods have not only better classification accuracy, but also show improved uncertainty estimation.
\end{abstract}
\begin{keywords}
automatic emotion recognition, emotion recognition in conversation, Dirichlet prior network, IEMOCAP
\end{keywords}
\section{Introduction}
\label{sec:intro}
Emotion understanding is a key attribute of conversational artificial intelligence (AI). 
Although significant progress has been made in developing deep-learning-based automatic emotion recognition (AER) systems over the past several years~\cite{tripathi2018multimodal,Majumder_2018,Poria2018,Han2018,Pappagari_2020}, most studies have focused on modelling and evaluating each utterance separately. However, emotions are known to be dependent on cross-utterance contextual information and persist across multiple utterances in dialogues~\cite{suls1998emotional}. 
This motivates the study of AER in conversation (ERC).

Emotion annotation is challenging due to the fact that emotion is inherently complex and ambiguous, and its expression and perception are highly personal and subjective. This causes uncertainty in the manual references used for emotional data. 
Although it is common to handle AER as a classification problem based on the majority agreed labels among several annotators \cite{Busso2008IEMOCAPIE, CHEAVD2.0,MSP-IMPROV,Meld}, it can cause two major problems. 
First, utterances without majority agreed labels have to be discarded, which makes the dialogue context non-contiguous in both training and test. 
Second, replacing the (possibly different) original labels from human annotators by the majority vote label also removes the inherent uncertainty associated with emotion perception in the data labelling procedure. 
To this end,   alternative methods to using the majority vote label are required for ERC.

Motivated by these problems, this paper proposes a novel distribution-based framework for ERC, which trains a dialogue-level Transformer model \cite{vaswani2017attention} to maximise the probability of generating a sequence of emotion distributions associated with a sequence of utterances. Each time step of the Transformer represents an utterance in the dialogue, and its corresponding input feature vector is the fusion of audio and text representations for that utterance derived using Wav2Vec 2.0 (W2V2) \cite{baevski2020wav2vec} and bidirectional encoder representations from Transformers (BERT) \cite{devlin-etal-2019-bert} respectively. 
The predicted emotion distribution of each utterance relies on all previous predictions as well as the audio and text features in the dialogue. 
By considering an emotion distribution as a continuous-valued categorical distribution, the original emotion class labels provided by the annotators can be viewed as samples drawn from the underlying true emotion distribution of the utterance. 
The proposed distribution-based ERC system then learns the true emotion distribution sequence in a conversation given the observed label samples. A novel training loss based on utterance-specific Dirichlet priors predicted by a Dirichlet prior network (DPN) is applied~\cite{wu2022estimating}, which improves distribution modelling performance by retaining the uncertainty in the original labels.  
Furthermore, by considering emotion as distributions, no utterances need to be discarded in either training or test, which keeps the dialogue context contiguous.

The rest of the paper is organised as follows. Section~\ref{sec: literature} provides background to ERC and the modelling of emotion ambiguity. Section~\ref{sec: proposed method} introduces distribution-based ERC. Section~\ref{sec: SSL} describes the use of representations derived by self-supervised learning (SSL) and the fusion of audio and text representations. The results and analysis are given in Section~\ref{sec: exp}, followed by conclusions.

\section{Related work}
\label{sec: literature}

\subsection{Emotion recognition in conversation}
\label{ssec: erc literature}
Emotion states can be described by discrete emotion categories (\textit{i.e.}, anger, happiness, neutral, sadness, \textit{etc.})~\cite{gunes2011emotion} or continuous emotion attributes (\textit{i.e.}, valence-arousal)~\cite{schlosberg1954three,nicolaou2011continuous}. This work focuses on classification-based AER using discrete emotion categories.
Much work has been published in classification-based AER using deep-learning-based methods~\cite{Majumder_2018,Han2018,Pappagari_2020,Yoon_2018}.
While good recognition performance has been achieved, the focus is on modelling information of each target utterance independently without considering the cross-utterance contextual information. 
Incorporating such information from both speakers in a dyadic conversation has been shown to improve AER performance~\cite{wu2021emotion}.

The conversational memory network (CMN)~\cite{hazarika2018conversational} was one of the first ERC approaches that used separate memory networks for both interlocutors participating in a dyadic conversation.
Built on the CMN, the interaction-aware attention network (IANN)~\cite{yeh2019interaction} integrates distinct memories of each speaker using an attention mechanism. 
Recently, the graph convolutional network was introduced to explore the relations between utterances in a dialogue~\cite{liu21o_interspeech,ghosal2019dialoguegcn,zhang2019modeling} where the representation of each utterance is treated as nodes and the relations between the utterances are the edges. 
Most of these models integrated a fixed-length context rather than the complete dialogue history.
Moreover, while contextual information was incorporated by including features of context utterances as inputs, the dependency of the output emotion states was not considered.

Several alternative structures have been proposed in response to the above two issues. The DialogRNN~\cite{majumder2019dialoguernn} approach uses a hierarchical recurrent neural network framework to recurrently model the emotion of the current utterance by considering the speaker states and the emotions of preceding utterances. The Transformer encoder structure has been adopted for ERC with a data augmentation method based on random utterance concatenation~\cite{pappagari2021beyond}. 
Another approach is to introduce an extra dialogue-level model on top of the single-utterance-based AER classifier. 
In the emotion interaction and transition method ~\cite{zhang2017interaction}, the emotion probability of each utterance is re-estimated using the previous utterance and currently estimated posteriors using an additional long short-term memory network. The
dialogical emotion decoding (DED) method~\cite{yeh2020dialogical} treats a dialogue as a sequence and consecutively decodes the emotion states of each utterance over time with a given recognition engine.

\subsection{Modelling emotion ambiguity}
The methods reviewed in Section~\ref{ssec: erc literature} only used utterances that have majority agreed emotion labels. 
However, emotion is inherently ambiguous and its perception is highly subjective, which leads to a large degree of uncertainty in labels. Different human annotators may assign different emotion labels to the same utterance and a considerable amount of data
does not have majority agreed labels from the human annotators. 
Majority voting results among the original labels are usually taken as the ground-truth label by AER datasets~\cite{Busso2008IEMOCAPIE, CHEAVD2.0,MSP-IMPROV,Meld}. Data without majority agreed labels are usually excluded from both training and test in classification-based AER, as they cannot be evaluated without ground-truth labels. In ERC, data exclusion can cause non-contiguous dialogue contexts for both training and test.
More importantly, the majority voting strategy considerably changes the uncertainty properties of the emotion states of the speaker~\cite{wu2022estimating}. Therefore, alternative methods for emotion state modelling are required. {Soft labels} have been used as targets in single-utterance-based AER~\cite{Ando_2018,Fayek_2016}, which averages the original {emotion class labels} provided by the annotators.  Such soft labels can be interpreted as the intensities of each emotion class and can allow 
all utterances to have a training label. Despite the use of soft labels, the systems were evaluated based on classification accuracy with the utterances with majority agreed labels, which results in a mismatch between the training loss and the evaluation metric ~\cite{Mower09interpretingambiguous}.
In this work, we propose a distribution-based method that allows the use of all utterances and all original labels for ERC.

\section{Distribution-based ERC}
\label{sec: proposed method}

\subsection{ERC as a sequence-to-sequence problem}

Consider a dialogue with $N$ utterances, let $\mathbf{x}_n$ and $\mathbf{y}_n$ be an utterance and its emotion state in terms of a probability distribution, ERC can be formulated as a special sequence-to-sequence 
problem, 
in which the input utterance sequence $\mathbf{x}_{1:N}$ and output emotion state sequence $\mathbf{y}_{1:N}$ have equal lengths and each output
distribution represents the emotion state of the corresponding input utterance.
Training can be performed by maximising $p(\mathbf{y}_{1:N}|\mathbf{x}_{1:N})$, the likelihood of generating the emotion states based on the input utterances. Based on the chain rule, $p(\mathbf{y}_{1:N}|\mathbf{x}_{1:N})$ can be calculated efficiently as: 
\begin{align}
\label{eq:eq1}
     p(\mathbf{y}_{1:N}|\mathbf{x}_{1:N})&=p(\mathbf{y}_{1}|\mathbf{x}_{1:N})\prod^{N}_{n=1}p(\mathbf{y}_{n+1}|\mathbf{y}_{1:n},\mathbf{x}_{1:N}).
\end{align}
Eqn.~\eqref{eq:eq1} differs from single-utterance-based AER \cite{tripathi2018multimodal,Majumder_2018,Poria2018,Han2018,Pappagari_2020} by conditioning $\mathbf{y}_{n+1}$ not only on $\mathbf{x}_{1:N}$ but also on $\mathbf{y}_{1:n}$, which reflects the fact that emotional states often persist across multiple utterances in a dialogue \cite{suls1998emotional}. 

\subsubsection{A Transformer for online ERC}
In Eqn.~\eqref{eq:eq1}, 
$p(\mathbf{y}_{n+1}|\mathbf{y}_{1:n},\mathbf{x}_{1:N})$ not only depends on the current and previous utterances $\mathbf{x}_{1:n+1}$, but also on future utterances $\mathbf{x}_{n+2:N}$, which is not suitable for online AER applications. Hence, an independence approximation between $\mathbf{y}_{n+1}$ and  $\mathbf{x}_{n+2:N}$ can be made:
\begin{align}
\label{eq:eq2}
     p(\mathbf{y}_{1:N}|\mathbf{x}_{1:N})&\approx p(\mathbf{y}_{1}|\mathbf{x}_{1})\prod^{N}_{n=1}p(\mathbf{y}_{n+1}|\mathbf{y}_{1:n},\mathbf{x}_{1:n+1}),
\end{align}
which is used throughout this paper. 

\begin{figure}[tb]
  \centering
  \includegraphics[width=\linewidth]{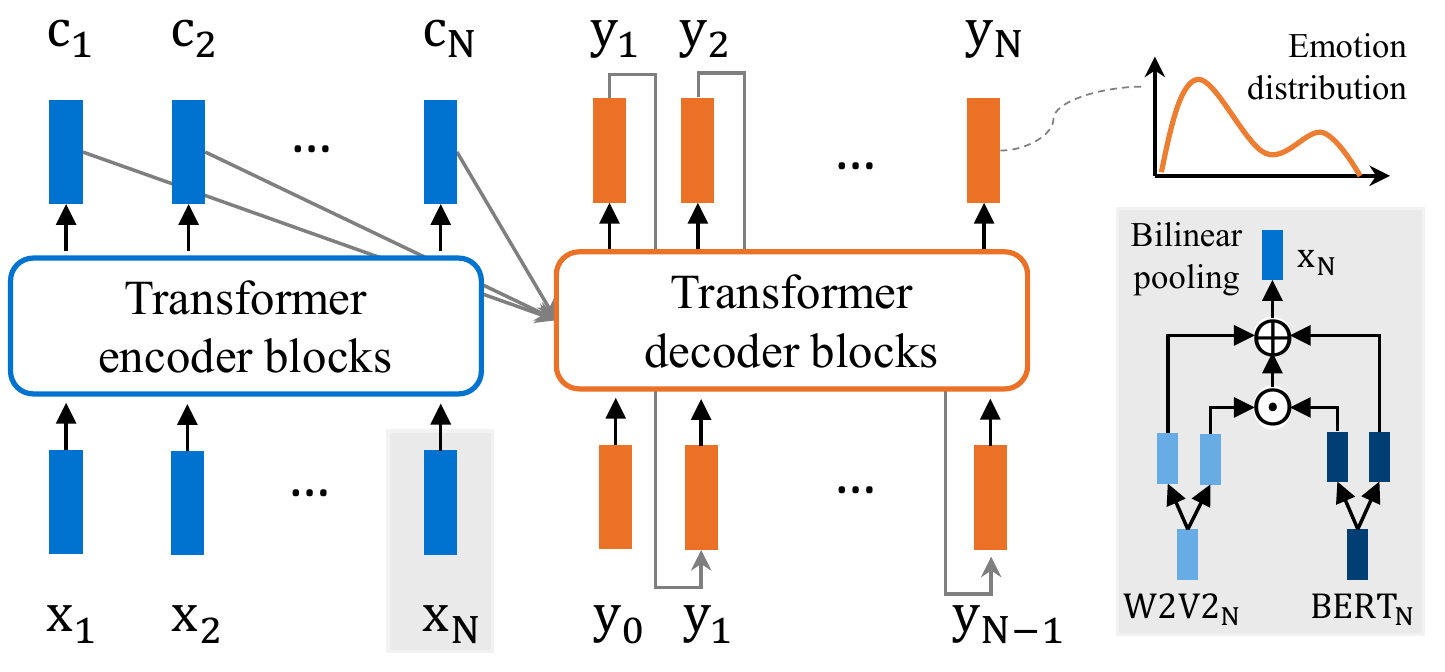}
  \caption{Schematic of the proposed distribution-based ERC system with a Transformer, where $\mathbf{x}_n$ is an utterance and $\mathbf{y}_n$ is the corresponding emotion distribution. Bilinear pooling is used to fuse the audio and text features derived from W2V2 and BERT.}
  \label{fig: structure}
\end{figure}

In this paper, Eqn.~\eqref{eq:eq2} is implemented with a Transformer encoder-decoder model~\cite{vaswani2017attention}, the schematic of the proposed system is shown in Fig.~\ref{fig: structure}. 
Teacher-forcing~\cite{williams1989learning} is commonly used when training an auto-regressive decoder structure where the output of the current time step depends on outputs of previous time steps. When making a prediction $\mathbf{y}_{n}$ at a time step $n$, teacher-forcing uses the ground-truth label history $\mathbf{t}_{1:n-1}$ during training, and uses the previous predictions output by the model $\mathbf{\hat{y}}_{1:n-1}$ during test. Teacher-forcing forces the decoder to over-fit to the ground-truth-label-based history and leads to a discrepancy between training and test.
Such a discrepancy can yield errors that propagate and accumulate quickly along the generated sequence.

\subsubsection{Avoid over-fitting to oracle history with scheduled sampling}

To alleviate the discrepancy caused by teacher-forcing, scheduled sampling~\cite{bengio2015scheduled} is used during training in this paper. A teacher-forcing ratio $\epsilon_i$ is introduced to randomly decide, during training, whether to use $\mathbf{t}_{n-1}$ or $\mathbf{\hat{y}}_{n-1}$:
\begin{equation}
	  \mathbf{{y}}_{n-1}=\left\{\begin{array}{ll}
      \mathbf{t}_{n-1}, & p_{\text{tf}} \leq \epsilon_i\\
	  \mathbf{\hat{y}}_{n-1}, & p_{\text{tf}} > \epsilon_i
	  \end{array}\right.
\end{equation}
where $p_{\text{tf}}$ is sampled from a uniform distribution between 0 and 1 ($p_{\text{tf}} \sim \mathbf{U}(0,1)$), $i$ denotes the $i^{th}$ mini-batch. $\epsilon_i$ gradually decreases as the training progresses, which changes the training process from a fully guided scheme based on previous ground-truth labels towards a less guided scheme based on previous model outputs. Commonly used schedules decrease $\epsilon_i$ as a function of $i$, which include a linear decay, an exponential decay and a inverse sigmoid decay~\cite{bengio2015scheduled}.

\subsection{Emotion distribution modelling using DPN}
\label{sec: dpn-kl}

Denote the underlying true emotion distribution of an utterance $\mathbf{x}$ as a categorical distribution $\boldsymbol{\mu}=\left[{p}(\omega_1|\boldsymbol{\mu}),\ldots,{p}(\omega_K|\boldsymbol{\mu})\right]^{\text T}$, where $K$ is the number of emotion classes. Emotion class labels $\omega_k$ from human annotators are samples drawn from this categorical distribution. Soft labels, as an approximation to the underlying true distribution, correspond to the maximum likelihood estimate (MLE) of $\boldsymbol{\mu}$ given the observed label samples $\{\boldsymbol{\mu}^{(1)},\ldots,\boldsymbol{\mu}^{(M)}\}$:
\begin{align}
\label{eqn: soft MLE}
    \bar{\boldsymbol{\mu}} &= \arg\max_{\boldsymbol{\mu}} \ln{p}(\boldsymbol{\mu}^{(1)},\ldots,\boldsymbol{\mu}^{(M)}|\boldsymbol{\mu})=\frac{1}{M}\sum_{m=1}^{M}\boldsymbol{\mu}^{(m)}.
\end{align}
Although soft labels enable contiguous dialogue contexts to be modelled by ERC, the MLE is a good approximation to the true  distribution only if a large number of original labels are available for each utterance, which cannot usually be satisfied in practice due to labelling difficulty and cost. Here, we use the Dirichlet prior network (DPN)~\cite{malinin2018predictive,Malinin2019ReverseKT,wu2022estimating} to resolve the label sparsity issue, which is a Bayesian approach modelling ${p}(\boldsymbol{\mu}|\boldsymbol{x})$ by predicting the parameters of its Dirichlet prior distribution.

\subsubsection{Emotion recognition with a Dirichlet prior}
The Dirichlet distribution, as the {conjugate prior} of the categorical distribution, is parameterised by its {concentration parameter} $\boldsymbol{{\alpha}}=[{\alpha}_1,\ldots,{\alpha}_K]^{\text T}$. A Dirichlet distribution $\operatorname{Dir}(\boldsymbol{\mu}| \boldsymbol{{\alpha}})$ is 
\begin{align}
    \label{eqn: dirchlet1} \operatorname{Dir}(\boldsymbol{\mu}| \boldsymbol{{\alpha}}) =&\frac{\Gamma\left({\alpha}_{0}\right)}{\prod_{k=1}^{K} \Gamma\left({\alpha}_{k}\right)} \prod_{k=1}^{K} \mu_{k}^{{\alpha}_{k}-1},\\ \nonumber{\alpha}_{0}=&\sum_{k=1}^{K} {\alpha}_{k},{~~}{\alpha}_{k}>0,
\end{align}
where $\Gamma(\cdot)$ is the gamma function defined as
\begin{align}
\Gamma\left({\alpha}_{k}\right)=\int^{\infty}_{0}z^{{\alpha}_{k}-1}e^{-z}\,dz.
\end{align}
In the Dirichlet process, given $\boldsymbol{{\alpha}}$, a categorical emotion distribution $\boldsymbol{\mu}$ is drawn from $\operatorname{Dir}(\boldsymbol{\mu}|\boldsymbol{{\alpha}})$, and an emotion class label $\omega_k$ is sampled from $\boldsymbol{\mu}$.

\subsubsection{DPN for emotion recognition}
A DPN is a neural network modelling $p(\boldsymbol{\mu}|\mathbf{x}, {\boldsymbol{\Lambda}})$ by predicting the concentration parameter ${\boldsymbol{\alpha}}$ of the Dirichlet prior, where $\boldsymbol{\Lambda}$ is the collection of model parameters. For each utterance $\mathbf{x}$, the DPN predicts ${\boldsymbol{\alpha}}=\operatorname{exp}[{f}_{{\boldsymbol{\Lambda}}}(\boldsymbol{x})]$, where $f_{{\boldsymbol{\Lambda}}}(.)$ is the DPN model. 
Note that the predicted ${\boldsymbol{\alpha}}$ depends on, and is specific to, each input utterance $\mathbf{x}$.
By predicting ${\boldsymbol{\alpha}}$ separately for each utterance, the DPN makes the Dirichlet prior ``utterance-specific'' and thus suitable for ERC tasks.

The predictive distribution of the DPN is the expected categorical distribution under $\operatorname{Dir}(\boldsymbol{\mu}| \boldsymbol{{\alpha}})$ \cite{malinin2018predictive}:
\begin{align}
\nonumber p(\omega_{k} | \mathbf{x}, {\boldsymbol{\Lambda}}) &=\mathbb{E}_{{p}(\boldsymbol{\mu} | \mathbf{x},{\boldsymbol{\Lambda}})}\left[{{p}}\left(\omega_{k} | \boldsymbol{\mu}\right)\right] \\
&= \frac{{{\alpha}_{k}}}{{\sum\nolimits_{k^{'}=1}^{K} \alpha_{k^{'}}}}=\operatorname{softmax}[{f}_{{\boldsymbol{\Lambda}}}(\boldsymbol{x})]_k,
\label{eqn: dpn pred}
\end{align}
which makes the DPN a normal neural network model with a softmax output activation function during test. 

DPN training is performed by maximising the likelihood of sampling the original labels (one-hot categorical distributions) from their relevant utterance-specific Dirichlet priors.
Given an utterance $\mathbf{x}$ with $M$ original labels $\{\boldsymbol{\mu}^{(1)},\ldots,\boldsymbol{\mu}^{(M)}\}$, a DPN is trained to minimise the negative log likelihood
\begin{align}
 \label{eqn: dpn loss}
  \mathcal{L}_{\text{dpn}}
  &=-\frac{1}{M}\sum\nolimits_{m=1}^{M} \ln {p}(\boldsymbol{\mu}^{(m)} | \mathbf{x},  \boldsymbol{\Lambda})\\
   \nonumber&=-\frac{1}{M}\sum\nolimits_{m=1}^{M}\ln \operatorname{Dir}(\boldsymbol{\mu}^{(m)} | f_{\boldsymbol{\Lambda}}(\mathbf{x})).
\end{align}
In contrast to soft labels that retain only the proportion of occurrences of each emotion class, the DPN preserves the information about each single occurrence of the emotion classes and also resolves the label sparsity issue with Dirichlet priors.

However, $\mathcal{L}_{\text{dpn}}$ is not directly applicable to an auto-regressive ERC system. This is because, in such a system, the targets of the previous time step are required for training (whether using teacher-forcing or scheduled sampling). In a DPN system, the output of the network is the hyperparameter $\alpha$ and the targets associated with $\alpha$ of the previous time step is unknown. Therefore, $\mathcal{L}_{\text{dpn}}$ was added as an extra term to the Kullback-Leibler (KL) divergence $\mathcal{L}_{\text{kl}}$ between the soft labels and the DPN predictive distributions. That is,
\begin{align}
\mathcal{L}_{\text{kl}}
&=\sum\nolimits_{k=1}^{K}-\bar{\mu}_k\ln{{p}(\omega_k|\boldsymbol{x},\boldsymbol{\Lambda})}+\sum\nolimits_{k=1}^{K} \bar{\mu}_k\ln{\bar{\mu}_k}
\label{eqn: soft kl}
\end{align}
\begin{align}
    \label{eqn: total loss}
    \mathcal{L}_{\text{dpn-kl}}&=\mathcal{L}_{\text{dpn}}+\lambda\,\mathcal{L}_{\text{kl}}.
\end{align}
The targets of the previous time step are the soft labels. This is a major difference from~\cite{wu2022estimating}, which treats the DPN loss as the main loss to model the uncertainty of emotions for each utterance independently without taking the previous emotion predictions into account.

\subsection{Evaluating distribution-based ERC}

Since classification accuracy cannot be applied to the utterances without majority agreed labels,  
it is no longer suitable to be the primary measure for evaluating distribution-based ERC system. The area under the precision-recall curve (AUPR) is used as an alternative metric~\cite{wu2022estimating} at test-time. A precision-recall (PR) curve is obtained by calculating the precision and recall for different decision thresholds where the $x$-axis of a PR curve is the recall, the $y$-axis is the precision. The AUPR is the average of precision across all recall values computed as the area under the PR curve. Compared to classification accuracy, AUPR can be applied to all test utterances and also quantify the model's ability to estimate uncertainty. 

In this paper, the curve is drawn by detecting utterances without majority agreed labels based on the model prediction. Utterances with majority agreed labels are selected as positive class and utterances without majority agreed labels are chosen as the negative class. Two threshold measures representing the confidence encapsulated in the prediction can be used as the threshold for AUPR:
\begin{itemize}
    \item The entropy of the predictive distribution (Ent.), defined as $-\sum\nolimits_{k=1}^{\mathrm{K}} {p}\left(\omega_{k} | \mathbf{x},\boldsymbol{\Lambda}\right)\ln {p}\left(\omega_{k}|\mathbf{x}, \boldsymbol{\Lambda}\right)$ that measures the flatness of the emotion distribution, where $\mathbf{x}$ is an utterance, $\omega_k$ is the $k$-th class and $\mathbf{\Lambda}$ is the model.

    \item The max probability (Max.P), $\max\nolimits_k {p}(\omega_{k}|\mathbf{x}, \mathbf{\Lambda})$ measuring the confidence of the predicted emotion class.
\end{itemize}

\section{SSL Representations for Audio and Text}
\label{sec: SSL}

Representations extracted from pre-trained universal models have recently attracted wide attention. These models are trained on a wide range of data at scale and can be fine-tuned to various downstream tasks and are sometimes referred to as foundation models~\cite{bommasani2021opportunities}. Self-supervised learning (SSL) is one of the most common approaches to train a foundation model as it does not require any labels from human annotators. It uses information extracted only from the input data itself in order to learn representations useful for downstream tasks, thus allowing the use of a large amount of unlabelled data for model training.

Models pre-trained by SSL have achieved great successes in natural language processing (\textit{e.g.} BERT~\cite{devlin-etal-2019-bert}, GPT‑2~\cite{radford2019language} and GPT-3~\cite{brown2020language}) and computer vision (\textit{e.g.}  ViT~\cite{dosovitskiy2020image} and iGPT~\cite{chen2020generative}), and have attracted increasing attention in speech processing \cite{mohamed2022self} (\textit{e.g.} ACPC~\cite{chorowski2021aligned}, W2V2~\cite{baevski2020wav2vec}, Hubert~\cite{hsu2021hubert}, and  WavLM~\cite{chen2022wavlm} \textit{etc}.). 
The large amount of unlabelled data leveraged by SSL can cover many linguistic and para-linguistic phenomena, and therefore could help to alleviate the data sparsity issue in AER \cite{makiuchi2021multimodal,morais2022speech,sharma2022multi,zou2022speech}. This paper used two SSL models: W2V2~\cite{baevski2020wav2vec} for the audio modality and BERT~\cite{devlin-etal-2019-bert} for the text modality\footnote{The proposed system can be viewed as a dialogue-level audio-text multimodal adaptor for emotions in the foundation model paradigm.}.

\subsection{Features derived from W2V2 and BERT}
\label{sec: feature}

W2V2 is a type of contrastive~\cite{oord2018representation} SSL model which learns representations by distinguishing a target sample (positive) from distractor samples (negatives) given an anchor representation. It takes as input a waveform and uses a convolutional feature encoder followed by a transformer network. This paper uses the ``wav2vec2-base" model\footnote{{https://huggingface.co/facebook/wav2vec2-base}} which contains 12 transformer blocks with model dimension 768, inner dimension 3,072 and 8 attention heads and is pre-trained using 960 hours of audio from Librispeech corpus~\cite{panayotov2015librispeech}. W2V2 representations were extracted from the output of the last Transformer block and averaged across each utterance.

BERT is a type of predictive SSL model which learns representations by predicting the masked tokens in a sentence. This paper uses the ``bert-base-uncased" model\footnote{{https://huggingface.co/bert-base-uncased}} which contains 12 transformer blocks with a model dimension of 768, inner dimension 3,072 and 12 attention heads and was pre-trained using a large amount of text.

\subsection{Bilinear-pooling-based feature fusion}
\label{sec: feature}
\begin{figure}[tb]
  \centering
  \includegraphics[width=0.85\linewidth]{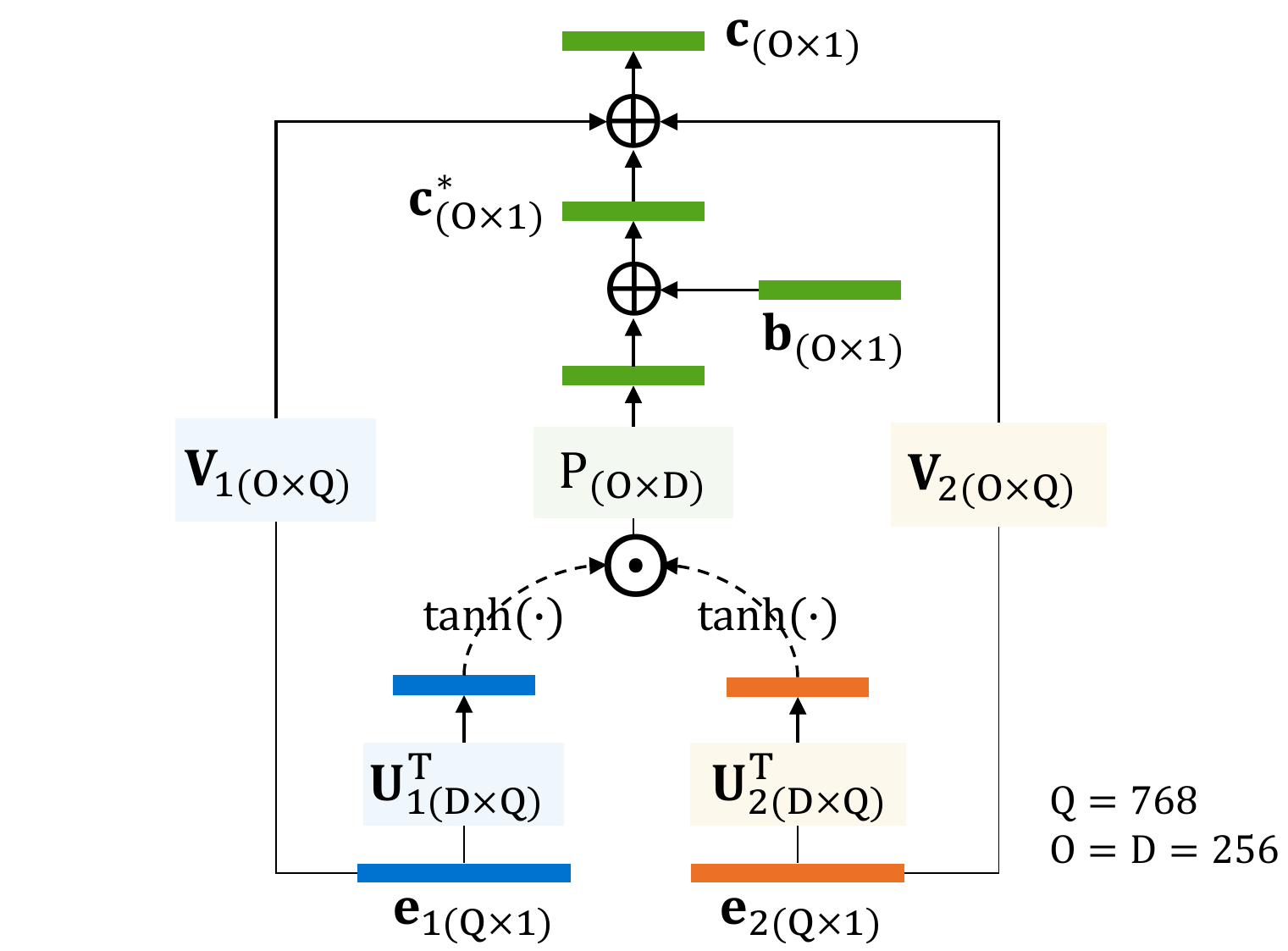}
  \caption{Schematic of bilinear pooling with shortcut combination. $\odot$ represents the Hadamard product and $\oplus$ represents the element-wise addition of vectors.}
  \label{fig:bilinear}
  \vspace{-2mm}
\end{figure}

Bilinear pooling~\cite{tenenbaum2000separating} is a commonly used approach for the expressive fusion of multimodal representations~\cite{zhang2020multimodal}, which models the multiplicative interactions between all possible element pairs. It computes the outer product of the $Q$-dimensional (-dim) audio and text representations, $\mathbf{e}_1$ and $\mathbf{e}_2$, into a $Q^2$-dim joint representation and then projects it into an $O$-dim space with a linear transform. 
In practice, bilinear pooling often suffers from a data sparsity issue caused by the high dimensionality of $Q^2$. Therefore, decomposition techniques are usually required in order to estimate the associated parameters properly and efficiently. 
In this paper, the W2V2 and BERT features were fused using a modified multimodal low-rank bilinear attention network with shortcut connections~\cite{sun2021combination}:
\begin{align*}
    \mathbf{c}^* = \mathbf{P}(\tanh(\mathbf{U}_1^{\text T} \mathbf{e}_1) \odot \tanh(\mathbf{U}_2^{\text T} \mathbf{e}_2)) + \mathbf{b}
\end{align*}
\begin{equation*}
        \mathbf{c} = \mathbf{c}^*+\mathbf{V}_1 \mathbf{e}_1 + \mathbf{V}_2 \mathbf{e}_2.
\end{equation*}
The process is illustrated in Fig.~\ref{fig:bilinear}. In this paper, $\mathbf{e}_1$, $\mathbf{e}_2$ are the W2V2 and BERT derived vectors to be combined, and are both 768-dim. $D$ and $O$ are both 256-dim, $\mathbf{c}$ is the 256-dim combined vector,  $\mathbf{U}_1$, $\mathbf{U}_2$ are both 768$\times$256-dim matrices, $\mathbf{b}$ is a 256-dim bias vector, $\mathbf{P}$ is a 256$\times$256-dim linear projection, $\mathbf{V}_1, \mathbf{V}_2$ are both 256$\times$768-dim, and $\odot$ is the Hadamard product. 

\section{Experiments}
\label{sec: exp}

\subsection{Experimental setup}
\subsubsection{Dataset}
The IEMOCAP~\cite{Busso2008IEMOCAPIE} corpus is used in this paper, which is one of the most widely used datasets for verbal emotion classification. It consists of 5 dyadic conversational sessions performed by 10 professional actors with a session being a conversation between two speakers. There are in total 151 dialogues including 10,039 utterances. Each utterance was annotated by three human annotators for emotion class labels (neutral, happy, sad, and angry \textit{etc.}). Each annotator was allowed to tag more than one emotion category for each sentence if they perceived a mixture of emotions, giving utterances 3.12 labels on average. Ground-truth labels were determined by majority vote 
Following prior work~\cite{tripathi2018multimodal,Majumder_2018,Poria2018}, the reference transcriptions from IEMOCAP were used for the text modality. Leave-one-session-out 5-fold cross validation (5-CV) was performed and the average results are reported.

\subsubsection{Data augmentation}
In the dialogue-level ERC system, the number of training samples is equal to the number of dialogues in the dataset, which is often very limited (\textit{e.g.} 151 in IEMOCAP). To mitigate training data sparsity issues, sub-sequence randomisation~\cite{li2021discriminative} was used to augment the training data, which samples sub-sequences $(\mathbf{x}_{s:e}, \mathbf{y}_{s:e})$ from each full training sequence as the augmented training samples, where $s$ and $e$ are the randomly selected start and end utterance indexes.

\subsubsection{Training specifications}
The Transformer architecture~\cite{vaswani2017attention} used for ERC contains 4 encoder blocks and 4 decoder blocks with a dimension of 256.
The multi-head attention contains 4 heads. Masking was applied to ensure that predictions for the current utterance depend only on previous input utterances and known outputs. Sinusoidal positional embeddings~\cite{vaswani2017attention} were added to the input features. A dropout rate of 10\% was applied to all parameters. 
The model was implemented using PyTorch. Scheduled sampling with an exponential scheduler was used during training. The Adam optimiser was used with a variable learning rate with linear warm-up for the first 2,000 training updates and then linearly decreasing.\footnote{Code availble: https://github.com/W-Wu/ERC-SLT22}

\subsection{AER with 4-way classification}
The most common setup on IEMOCAP~\cite{Majumder_2018,Han2018,Yoon_2018, yeh2019interaction,yeh2020dialogical,liu21o_interspeech} only uses utterances with majority agreed labels belonging to ``angry", ``happy", ``excited" (merged with ``happy''), ``sad", and ``neutral" for a 4-way classification, which results in 5,531 utterances. For comparison, 4-way emotion classification systems using this setup were first built as baselines.
Since the test sets are slightly imbalanced between different emotion categories, both weighted accuracy (WA) and unweighted accuracy (UA) are reported for classification experiments. WA corresponds to the overall accuracy while UA corresponds to the mean of class-wise accuracy.

The ``wav2vec2-base" model was fine-tuned for single-utterance-based 4-way classification by adding a 128-d fully connected layer and an output layer with softmax activation on top of the pre-trained model. The fine-tuning experiment results are shown as ``utterance-W2V2ft" in Table~\ref{tab: classification}. The Transformer ERC model was then trained using the
fine-tuned W2V2 features (``dialogue-W2V2ft"), BERT features (``dialogue-BERT"), and the fusion of BERT and the fine-tuned W2V2 features (``dialogue-W2V2ft+BERT") as input.
Comparing ``utterance-W2V2ft" and ``dialogue-W2V2ft" in Table~\ref{tab: classification}, dialogue-based AER performs better than single-utterance-based AER. The fusion of audio and text features further improves the performance.
\begin{table}[tb]
\centering
\caption{5-fold CV classification results (mean$\pm$ standard deviation across folds) for 4-way utterance and dialogue baseline systems on IEMOCAP. IANN~\cite{yeh2019interaction} and DED~\cite{yeh2020dialogical} are ERC methods using audio features only without pre-trained encoders.} 
\vspace{1mm}
\begin{tabular}{l|cc}
\toprule
 System &  {\%WA}  & {\%UA}\\
\midrule
utterance-W2V2ft & 68.71$\pm$2.60        & 69.99$\pm$3.91  \\
\midrule
dialogue-W2V2ft & 70.18$\pm$4.83        & 71.32$\pm$4.06\\
dialogue-BERT & 68.57$\pm$3.50       & 67.56$\pm$2.94   \\
dialogue-W2V2ft+BERT &  74.87$\pm$3.77           & 74.59$\pm$3.50   \\
\midrule
\midrule
IANN~\cite{yeh2019interaction} &64.7 & 66.3\\
DED~\cite{yeh2020dialogical} & 69.0 & 70.1 \\
\bottomrule
\end{tabular}
\label{tab: classification}
\vspace{-3mm}
\end{table}

\subsection{IEMOCAP data analysis and motivation for distribution-based systems}

\begin{figure}[]
  \centering
  \includegraphics[width=\linewidth]{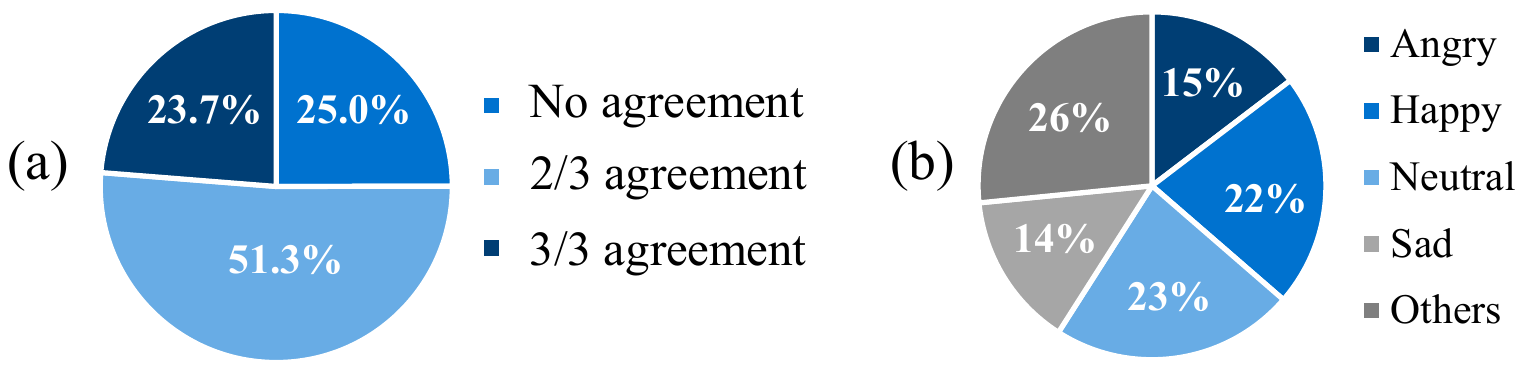}
  \caption{Distribution of data in IEMOCAP. (a) Proportion of annotators agreeing on the label. (b) Ground-truth of utterances with unique majority labels.}
  \vspace{-1ex}
  \label{fig:stats}
\end{figure}

Statistics for the IEMOCAP corpus are summarized in Fig.~\ref{fig:stats}. The 4-way classification setup discards 45\% of the  data in IEMOCAP that belongs to the following two categories:
\begin{itemize}
    \item Utterances without majority agreed emotion labels (2,507 utterances);
    \item Utterances whose majority agreed labels do not belong to the selected four emotion classes in the 4-way setup (2,001 utterances).
\end{itemize}
This strategy not only causes a loss of nearly half of the emotion data, which are highly valuable, but also causes the dialogue contexts modelled by the Transformer to be non-contiguous.
Furthermore, among the utterances with majority agreed labels, only 31.6\% have all annotators agreed on the same emotion class label. When majority voting is applied, an utterance with labels ``happy", ``happy", ``happy" and an utterance with labels ``happy", ``happy", ``sad" have the same ground-truth label ``happy", even though an annotator has assigned a different label to the latter utterance. The use of majority voting therefore changes the true emotion distributions and causes the inherent uncertainty that exists in the annotations to be discarded.

\begin{figure*}[]
  \centering
   \includegraphics[width=\linewidth]{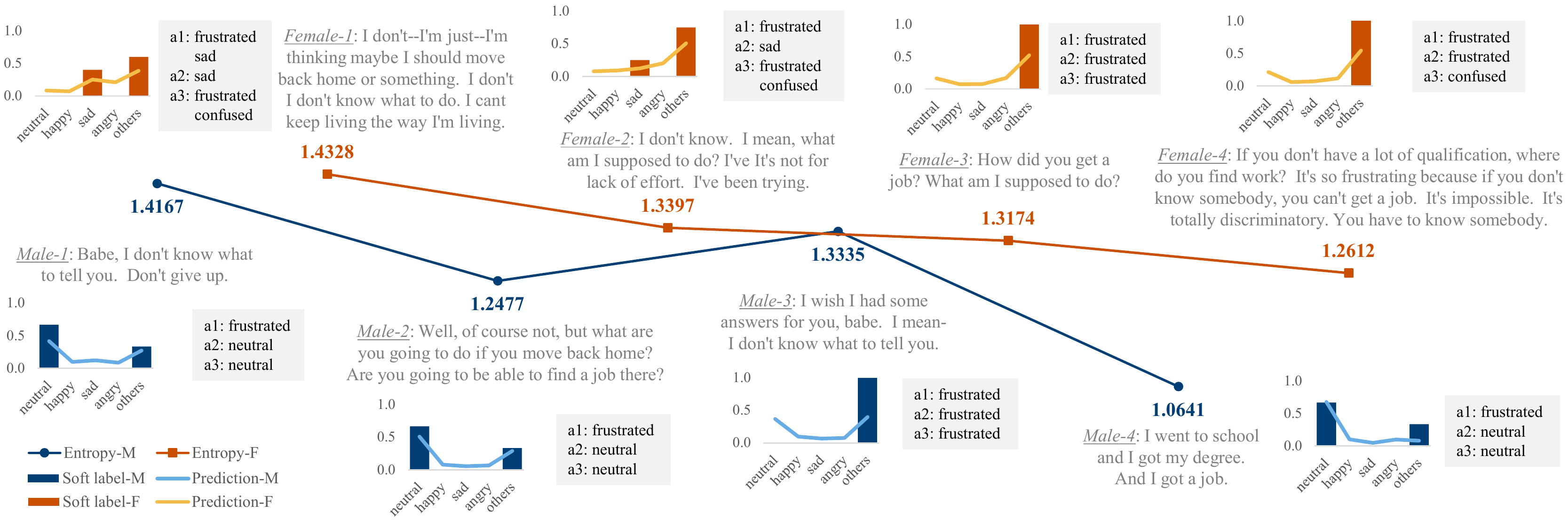}
   \vspace{-3ex}
  \caption{Entropy of the predicted emotion distribution of each utterance in a sub-dialogue. The DPN-KL system  trained on Session 1-4 was used. For each sentence, the bar chart shows the soft label and the line on the bar chart shows the prediction. Labels provided by the three annotators are shown in the grey box, with ``a1" referring to the first annotator \textit{etc}. (``frustrated" and ``confused" are merged into the 5-th class ``others").}
  \label{fig:case}
  \vspace{-2ex}
\end{figure*}

\subsection{ERC with Distribution modelling}

In this section, the training targets for each utterance were revised from a single discrete label to a continuous-valued {categorical distribution} over five emotion classes. The five classes correspond to the previous four emotion classes plus an extra class ``others" that includes all the other emotions that exist in IEMOCAP.
This not only allows all data in IEMOCAP to be used for ERC, regardless of whether majority agreed label exists and whether it belongs to the 4 classes, but also avoids the problem that not all of the original labels are represented by majority voting. 
Three systems were evaluated:
\begin{itemize}
    \item HARD: A system trained by 5-way classification. When training the HARD system, all utterances in a dialogue were taken as input while the loss was only computed for utterances that have majority agreed labels.
    \item SOFT: A KL-only system trained by minimising $\mathcal{L}_{\text{kl}}$ in Eqn.~\eqref{eqn: soft kl}.
    \item DPN-KL: A system trained by minimising the combined loss $\mathcal{L}_{\text{dpn-kl}}$ in Eqn.~\eqref{eqn: total loss} with $\lambda=20$.
\end{itemize}
All systems were first evaluated by 5-way classification accuracy on test utterances with majority agreed labels, as well as by 4-way classification accuracy on test utterances whose majority agreed labels belong to the 4 classes,
and then evaluated by the average AUPR (Max.P) and AUPR (Ent.) on all test utterances.

\begin{table}[htb]
\centering
\caption{5-fold CV results for the proposed distribution-based ERC method on IEMOCAP. Highest value in each row shown in bold font.} 
\begin{tabular}{c|ccc}
\toprule
 Metric  & {HARD}  & {SOFT}  & {DPN-KL} \\
\midrule
5-way \%WA & 59.99$\pm$4.14  & 62.48$\pm$3.70  & \textbf{63.63}$\pm$2.43  \\
5-way \%UA & 58.83$\pm$3.51  & {62.54}$\pm$3.98  & \textbf{63.12}$\pm$3.30  \\
\midrule
4-way \%WA & 73.01$\pm$3.79  & {77.46}$\pm$2.49  & \textbf{77.83}$\pm$2.07  \\
4-way \%UA & 72.57$\pm$3.23  & \textbf{78.16}$\pm$2.88  & {78.12}$\pm$2.60  \\
\midrule
\%AUPR (MaxP) & 76.63$\pm$2.09 & 78.02$\pm$1.32 & \textbf{80.72}$\pm$1.67 \\
\%AUPR (Ent.)  & 76.99$\pm$2.24 & 78.63$\pm$1.27 & \textbf{81.17}$\pm$2.02  \\
\bottomrule
\end{tabular}
\label{tab: distribution modelling}
\end{table}

\begin{figure}[tb]
  \centering
  \includegraphics[width=0.49\linewidth]{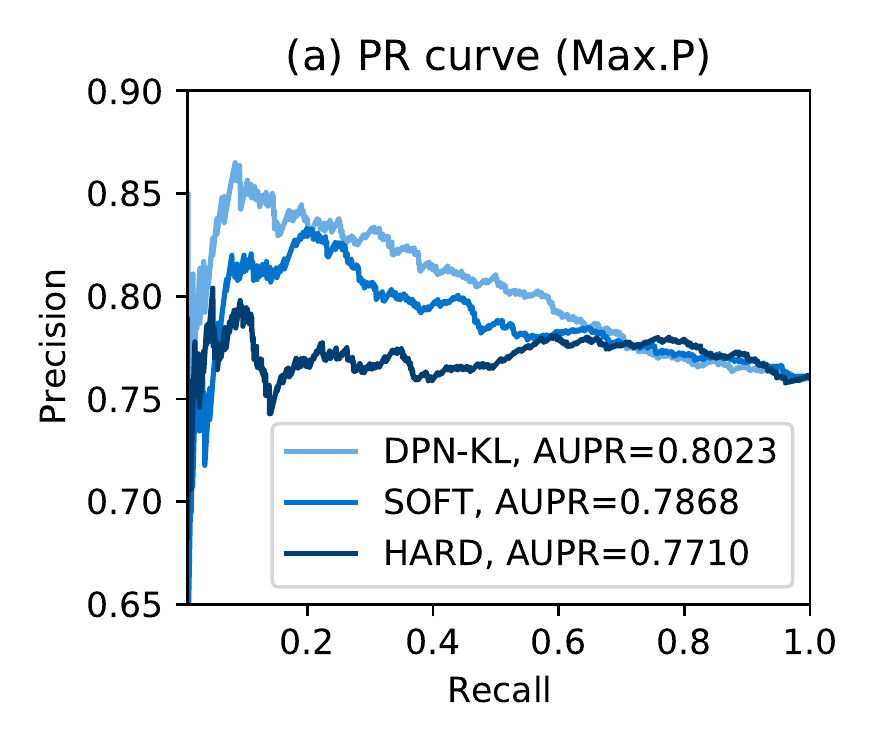}
  \includegraphics[width=0.49\linewidth]{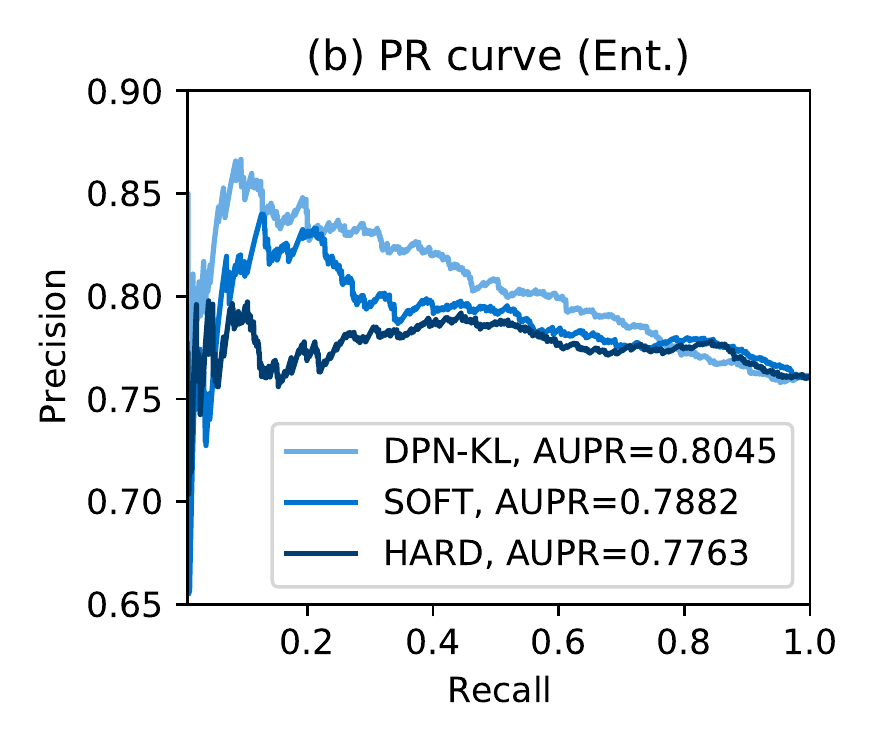}
  \vspace{-4mm}
  \caption{PR curves of the three systems using (a) Max.P and (b) Ent. as the uncertainty measures. The tests were performed on Session 5.}
  \vspace{-5mm}
  \label{fig:AUPR}
\end{figure}

As shown by the results in Table~\ref{tab: distribution modelling}, both the SOFT and DPN-KL systems outperform the HARD system on all evaluation metrics. 
A possible explanation lies in the fact that emotion assignment errors in the hard classification setup (HARD system) are more likely to propagate through the dialogue with the auto-regressive Transformer decoder. 
The DPN-KL system produces the highest AUPR among all of the systems, which shows that it has the best performance in modelling emotion distributions as it gives the best prediction of utterances without majority agreed labels. For the convenience of visualisation, the 5th fold (trained on Session~1-4 and tested on Session~5) was taken as an example and the PR curves of test utterances for all three systems are shown in Fig.~\ref{fig:AUPR}. It can be seen that the DPN-KL system gives the largest area under the PR curve, showing its superior uncertainty estimation performance.

\subsection{Analysis}
This section gives a case study to better understand uncertainty variation of emotion distributions in a dialogue. Fig.~\ref{fig:case} shows the trend of uncertainty change in emotion estimation (measured by the entropy of the predicted emotion distribution) in a sub-dialogue selected from the dialogue ``Ses05F\_impro04" in IEMOCAP Session 5 between a female and a male speaker. The results were produced by the DPN-KL system trained on Session 1-4. For each sentence, the bar chart shows the soft label and the line on the bar chart shows the prediction. Labels provided by the three annotators are shown in the grey box.

From Fig.~\ref{fig:case}, utterance \textit{Female-1} has two annotators that each provided two labels, indicating the uncertainty of the emotional content of the utterance. The uncertainty of emotion estimation is reflected by the high entropy. Although utterances \textit{Female-2} and \textit{Female-3} have the same emotion class labels found by majority voting, their underlying emotion distributions are different. As the dialogue progressed (from \textit{Female-1} to \textit{Female-3}), the annotators gradually became more certain that the female speaker got frustrated (shown by the soft labels), and the predicted distribution changed accordingly. Given the same label samples (\textit{i.e.,} \textit{Female-3} and \textit{Female-4}; \textit{Male-1}, \textit{Male-2} and \textit{Male-4}), the entropy decreases as the dialogue progressed, indicating that the model is becoming more confident about its predictions.
The reductions of uncertainty in this example demonstrates the advantage of using cross-utterance contextual information in our proposed distribution-based ERC framework. Moreover, another example is the emotional shift that occurs from utterance \textit{Male-2} to \textit{Male-3}. Due to emotional inertia, the model predicts a higher probability of ``others" while still retaining some probability for ``neutral". The larger entropy reveals that the model is uncertain about this prediction.

\section{Conclusion}
In this paper, we propose a distribution-based ERC framework, which formulates ERC as a special sequence-to-sequence problem for distribution estimation. The emotion state of an utterance is represented by a categorical distribution which depends on the context information and emotion states of the previous utterances in the dialogue. Each input vector of a dialogue sequence input to the Transformer dialogue model is formed by fusing representations extracted from SSL models W2V2 and BERT, which makes the system also a dialogue-level audio-text multimodal task adaptor for AER. The system is trained by minimising the KL divergence combined with the DPN loss which conditions the categorical distribution on an utterance-specific Dirichlet prior distribution, which is evaluated by AUPR with the task of detecting utterances without majority agreed labels. This approach not only allows utterances without majority agreed labels to be used, but also leads to better performance in modelling the uncertainty variations in ERC.

\bibliographystyle{IEEEbib}
\bibliography{strings,refs}

\end{document}